\documentclass[10pt,twocolumn,letterpaper]{article}

\usepackage[pagenumbers]{iccv} 

\usepackage{amsmath}
\usepackage{autobreak}
\usepackage{graphicx}
\usepackage{multirow}
\definecolor{iccvblue}{rgb}{0.21,0.49,0.74}
\usepackage[pagebackref,breaklinks,colorlinks,allcolors=iccvblue]{hyperref}

\def\qi{\textcolor{black}}

\title{Collaborative Temporal Consistency Learning for Point-supervised Natural Language Video Localization}

\author{Zhuo Tao$^{1,2}$~~~Liang Li$^{1}$~~~Qi Chen$^{5}$~~~Yunbin Tu$^{2}$~~~Zheng-Jun Zha$^{6}$~~~Ming-Hsuan Yang$^{4}$\\~~~Yuankai Qi$^{3}$~~~Qingming Huang$^{1,2}$\\
	$^1$Institute of Computing Technology, Chinese Academy of Sciences~~~$^3$Macquarie University\\
    $^2$University of Chinese Academy of Sciences~~~$^4$University of California at Merced~~~\\$^5$University of Adelaide~~~$^6$University of Science and Technology of China\\
    }

\begin{document}
\maketitle
\begin{abstract}
Natural language video localization (NLVL) is a crucial task in video understanding that aims to localize the target moment in videos specified by a given language description. 
Recently, a point-supervised paradigm has been presented to address this task, requiring only a single annotated frame within the target moment rather than complete temporal boundaries.  Compared with the fully-supervised paradigm, it offers a balance between localization accuracy and annotation cost. 
However, due to the absence of complete annotation, it is challenging to align the video content with language descriptions,
consequently hindering accurate moment prediction.
To address this problem, we propose a new \textbf{C}\textbf{O}llaborative \textbf{T}emporal consist\textbf{E}ncy \textbf{L}earning (COTEL) framework that leverages the synergy between saliency detection and moment localization to strengthen the video-language alignment.
Specifically, we first design a frame- and a segment-level Temporal Consistency Learning (TCL) module that models semantic alignment across frame saliencies and sentence-moment pairs.
Then, we design a cross-consistency guidance scheme, including a Frame-level Consistency Guidance (FCG) and a Segment-level Consistency Guidance (SCG), that enables the two temporal consistency learning paths to reinforce each other mutually.
Further, we introduce a Hierarchical Contrastive Alignment Loss (HCAL) to comprehensively align the video and text query.
Extensive experiments on two benchmarks demonstrate that our method performs favorably against SoTA approaches.
We will release all the source codes.
\end{abstract}    
\section{Introduction}
\label{sec:intro}
\begin{figure}
    \centering
    \includegraphics[width=\linewidth]{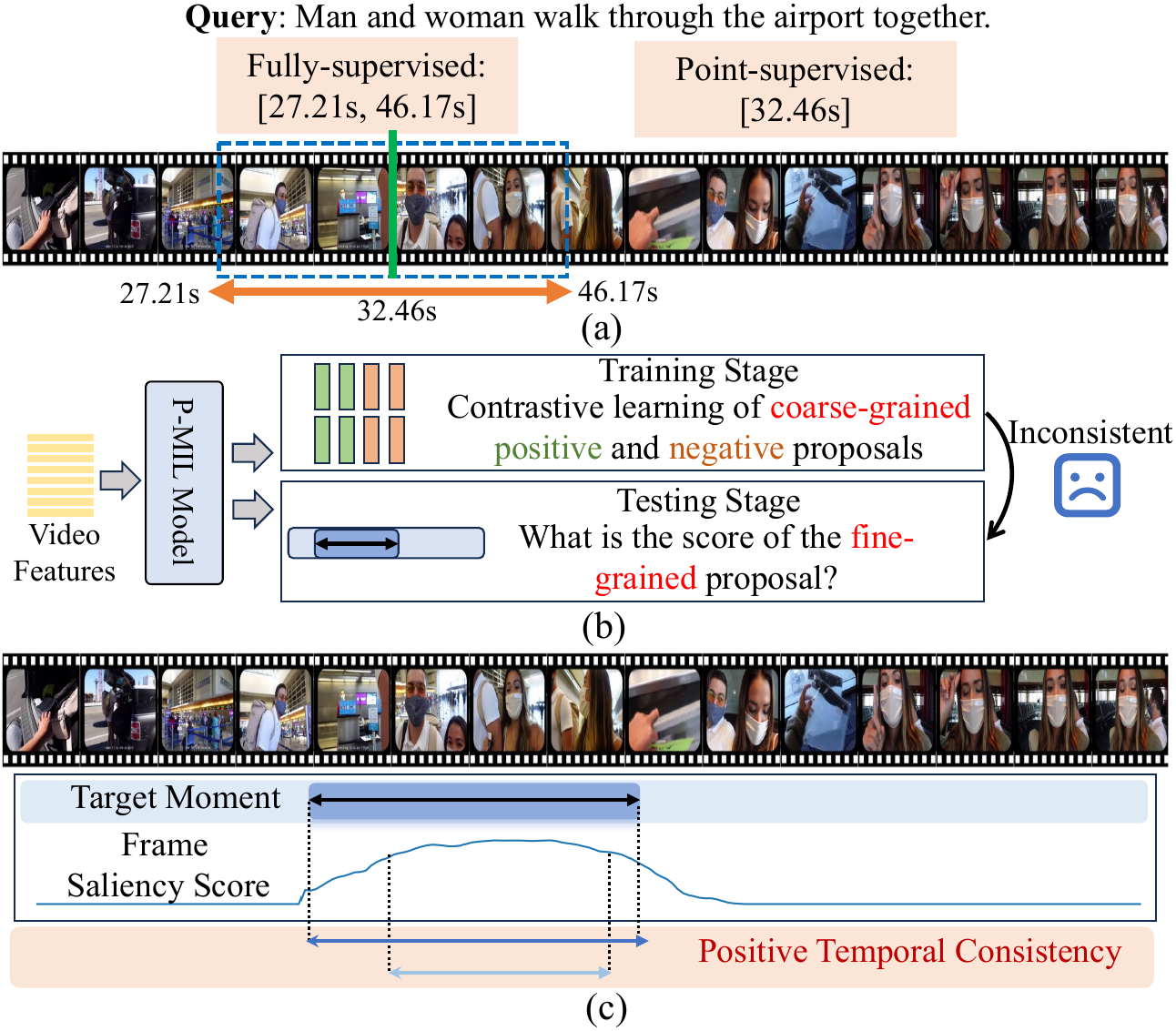}
    \caption{(a) Different supervision signals. (b) The proposals in the training stage are coarse-grained while fine-grained in the test stage, leading to inconsistency. (c) Frame saliency detection and moment localization share an extensive temporal correspondence between their respective positive temporal segments.}
    \label{fig:introduction}
    \vspace{-1em}
\end{figure}

Natural language video localization (NLVL), a fundamental task in video understanding, has gained increasing attention in recent years. NLVL requires the model to localize the target segment in the video that is semantically related to the given query.

Previous methods~\cite{2D-TAN,MCN,CTRL} typically solve this task under a fully-supervised setting, as shown in 
Fig.~\ref{fig:introduction}~(a), where the start and end timestamps of the target moment are needed. 
However, obtaining such complete annotations is time-consuming and labor-intensive, which severely hinders the application of NLVL in real-world scenarios. 
Besides, Otani \textit{et al.}~\cite{Otani_Nakashima_Rahtu_Heikkilä_2020} show that the annotated temporal boundaries may vary when different annotators process the same query-video pairs. These unavoidable inconsistencies in manual annotations can potentially compromise the model's ability to achieve visual-text alignment. 

Recently, Cui \textit{et al.}~\cite{viga} propose an annotation paradigm, termed point-supervised annotation, which requires a single timestamp within the temporal boundary of the target moment. 
Although its performance falls short of fully-supervised methods, it is about six times more cost-effective \cite{lee2021learning}, offering a favorable balance between annotation efficiency and model performance. 
However, the main challenge in this paradigm lies in the mismatch between the supervision signal from a single frame and the segment that needs to be predicted. 
To address this challenge, prior methods~\cite{viga,d3g} typically adopt the proposal-based multi-instance learning strategy (P-MIL). Specifically, they use sliding windows~\cite{viga} or 2D maps~\cite{d3g} to split the video into multiple segment proposals,
which are assigned with Gaussian weights according to the point annotation. 
They treat proposals containing the annotation point as positive instances and those without as negative instances, aligning the video and text query through contrastive learning.

Despite the progress,
their performance is still limited as their segment proposals are coarse-grained due to the sampling strategy, 
which are not well aligned with the real target moment.
Therefore, as shown in Fig.~\ref{fig:introduction}~(b), the model struggles to accurately predict the score for target moment.
However, we observe that frames in the target moments tend to have higher saliency scores, showing a strong correlation between moment localization and frame saliency detection. 
As shown in Fig.~\ref{fig:introduction}~(c), these two types of predictions share an extensive temporal correspondence between their respective positive temporal segments, showing significant consistencies. 
Motivated by this observation, we propose to leverage the synergy between saliency detection and moment localization to strengthen the alignment between video and query.

Based on the above observation, we propose a novel \textbf{C}\textbf{O}llaborative \textbf{T}emporal consist\textbf{E}ncy \textbf{L}earning (COTEL) framework that integrates both frame- and segment-level temporal consistency learning (TCL). 
In this framework, we design a frame-level TCL to detect the saliency of frame sequences in a 1D temporal fashion based on Gaussian prior.
We also incorporate a segment-level TCL to localize target moments based on P-MIL~\cite{viga,d3g} strategy.
With these two branches, we devise a cross-consistency guidance mechanism, consisting of a frame-level consistency guidance (FCG) module and a segment-level consistency guidance (SCG) module, to learn mutual synergy between two types of TCL.
In FCG, saliency scores from frame-level TCL are first processed through a saliency-aware mask generator to generate a saliency-aware consistency mask, which is then used to enhance video-text features via element-wise product for the segment-level TCL.
Similarly, in SCG, proposal scores from segment-level TCL are transformed into a semantic-aware consistency mask through a semantic-aware mask generator, which enhances the video-text features for frame-level TCL through element-wise multiplication. 
\qi{Last, to enhance video-text alignment in segment-level TCL, we propose a hierarchical contrastive alignment loss (HCAL) with two components: 1) intra-video selective alignment, which aligns $k$ key positive samples to improve semantic understanding and boundary perception, and 2) inter-video contrastive mining, which distinguishes negative samples across videos for robust contrastive learning.}

The main contributions of this paper are as follows:
\begin{itemize}
    \item We propose a novel COTEL framework that integrates both frame-level and segment-level temporal consistency learning (TCL) for point-supervised NLVL, which learns the synergy between saliency detection and moment localization, thereby enhancing the cross-modal alignment between video and query.
    \item We design a hierarchical contrastive alignment loss (HCAL) to align the features of the positive sentence-moment pairs and the cross-consistency guidance to leverage the temporal consistency between saliency detection and moment localization to ensure better alignment between video and query.
    \item Extensive experiments on two datasets demonstrate the effectiveness and favorable performance of the proposed method compared to several state-of-the-art approaches.
\end{itemize}


\section{Related Work}
\label{sec:related work}

\paragraph{Fully-Supervised Natural Language Video Localization.}
There are mainly two paradigms for fully supervised natural language video localization: proposal-based and proposal-free methods. (i) Proposal-based methods~\cite{ACRN, TSGVpaperreviewzongshu, TSGVpaperreviewzongshu2} can be divided into three types based on how proposal candidates are generated: sliding window-based~\cite{ACRN,CTRL,MCN}, proposal-generated~\cite{QSPN,SAP,SPN}, and anchor-based methods~\cite{TGN, MAN, SCDM}. Sliding window-based and certain proposal-generated methods use a two-stage propose-and-rank pipeline in which proposal candidates are generated apart from model computations. They first generate the proposal, then they rank them based on the similarity between the proposals and the input sentence. Anchor-based approaches combine proposal creation into model computation, resulting in end-to-end learning.  (ii) Proposal-free methods~\cite{ablr,excl,vslnet} directly predict the start and end boundaries of the target moments on fine-grained video snippet sequence.  Proposal-free approaches are classified as regression-based~\cite{ablr} or span-based~\cite{vslnet}, depending on the format of moment boundaries. Although fully-supervised methods have achieved satisfactory performance, they are highly dependent on accurate timestamp annotations. However, obtaining these annotations for large-scale video-sentence pairs is extremely time-consuming and labor-intensive, making it difficult to be used in real-world scenarios. Furthermore, annotations suffer from the incorrect issue, as action boundaries in videos are typically subjective and variable across annotators.

\vspace{-1em}
\paragraph{Point-Supervised Natural Language Video Localization.}
Recently, Cui~\etal~\cite{viga} propose a new point-supervised paradigm. This paradigm requires the timestamp of only one single random frame, which we refer to as a ``point", within the temporal boundary of the fully supervised counterpart. ViGA~\cite{viga} utilizes a Gaussian function to model the relevance of different proposals with target moment and contrasts the proposals with the queries. D3G~\cite{d3g} generates 2d map proposals and dynamically adjusts the distribution to reduce the annotation bias. DBFS~\cite{yang2023probability} introduces distribution functions to model both the probability of the action frame and that of the boundary frame. CFMR~\cite{cfmr} introduces a concept-based multimodal alignment mechanism to achieve efficient and cost-effective video moment retrieval, significantly improving retrieval speed while maintaining high accuracy. Despite great success, these methods suffer from the discrepancy of single-frame annotation in the training stage and predicting the target moment in the test stage. The coarse-grained proposal during training is hard to cover the fine-grained target moments during inference. This inconsistency hinders the alignment learning of the target moment and the input query, thus leading to suboptimal performance.

\section{Methodology}

\begin{figure*}[h]
    \centering
    \includegraphics[width=\linewidth]{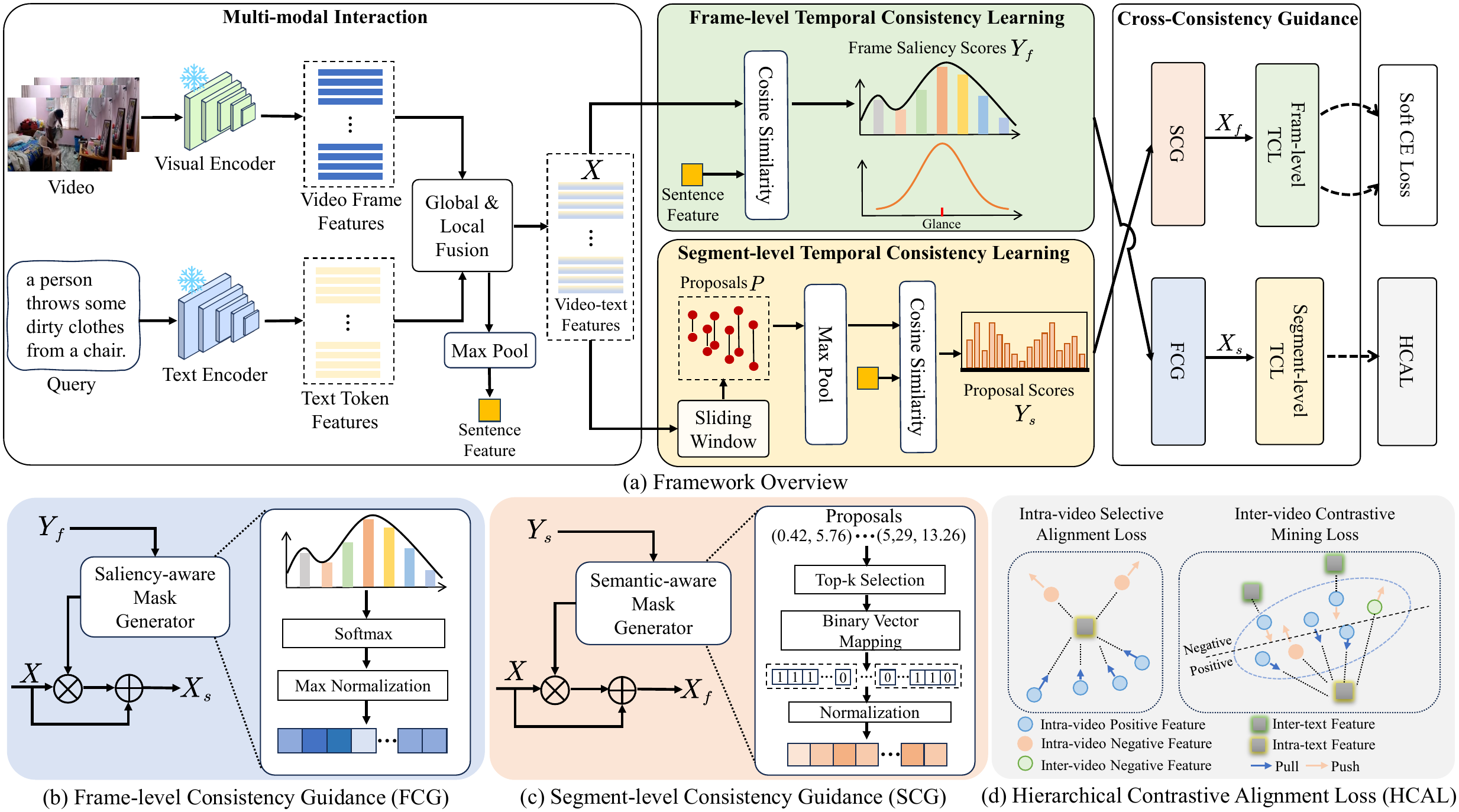}
    \caption{(a) Overview of our collaborative temporal consistency learning framework, which consists of multi-modal interaction (Sec.\ref{sec:Multi-modal Interaction}), frame-level temporal consistency learning (Frame-level TCL, Sec.\ref{sec:Frame-level Temporal Consistency Learning}), segment-level temporal consistency learning (Segment-level TCL, Sec.\ref{sec:Segment-level Temporal Consistency Learning}) and cross-consistency guidance (Sec.\ref{sec:cross consistency guidance}). (b) The frame-level consistency guidance (Sec.\ref{sec:Frame Level Guidance}) utilizes the frame-level saliency scores to enhance the fine-grained alignment of video and text in the segment-level moment localization. (c) Segment-level consistency guidance (Sec.\ref{sec:Segment Level Guidance}) generates the semantic-aware consistency mask with a proposal-based mask generator to guide frame-level TCL. (d) Hierarchical contrastive alignment loss (HCAL, Sec.\ref{sec:Hierarchical Contrastive Alignment Loss}) consists of intra-video selective alignment and inter-video contrastive mining losses to regularize the alignment between the video and paired sentence.}
    \label{fig:model}
    \vspace{-1em}
\end{figure*}

\subsection{Overview}

\paragraph{Problem Definition.}
\label{sec:Problem Definition}
Given an untrimmed video and a language query, the task of point-supervised natural language video localization seeks to determine the temporal boundaries of the target moment, represented as ($\tau_s,\tau_e$), where $\tau_s$ and $\tau_e$ denote the start and end frames related to the given query. 
There is solely one annotated frame $t_p$ for a given query during training, where $\tau_s \leq t_p \leq \tau_e$.

\paragraph{Pipeline.}
In the absence of boundary annotations, we propose a framework that leverages the synergy between frame-level and segment-level temporal consistency learning (TCL) for point-supervised NLVL, as shown in Fig.~\ref{fig:model}~(a). Our framework comprises three main steps:
First, given a video-query pair, we extract features using pre-trained video and text encoders, then fuse these features through a multi-modal interaction module.
Second, we process the video-text features through both frame-level and segment-level TCL branches to perform video-text alignment at complementary granularities.
Third, to learn mutual synergy between the two types of TCL, we introduce the cross-consistency guidance consisting of frame-level consistency guidance (FCG) and segment-level consistency guidance (SCG). In FCG, saliency scores from frame-level TCL generate a saliency-aware consistency mask that enhances video-text features for segment-level TCL via element-wise multiplication. Similarly, in SCG, proposal scores from segment-level TCL create a semantic-aware consistency mask that enhances video-text features for frame-level TCL. The enhanced features are then reprocessed through both TCLs to produce the final frame saliency scores and segment proposal scores. We employ a soft cross-entropy (CE) loss to train frame-level TCL and design a hierarchical contrastive alignment loss (HCAL) to optimize the segment-level TCL for better video-text alignment. Below we provide details of each main component.

\subsection{Multi-modal Interaction}
\label{sec:Multi-modal Interaction}

We first use pre-trained CNN (e.g., C3D~\cite{C3D}, I3D~\cite{I3D}) to extract video features to ensure a fair comparison. Then we uniformly sample $L_v$ features and project them into $d$ dimensional representations using a fully connected layer. We also incorporate fixed positional encoding into the model and get enhanced visual representation $V \in \mathbb{R}^{L_v \times d}$, where $d$ represents the dimension of the visual representation.
Given an input language query, following ViGA~\cite{viga}, we start by initializing its word features using GloVe embeddings~\cite{glove}, followed by a fully connected layer to project their dimensions to $d$. Analogous to the visual representation, positional encoding is also added to the text representation, resulting in the enhanced features $Q \in \mathbb{R}^{L_q \times d}$. 

After encoding the video and language query, we employ two multi-head cross-attention (MHCA) layers \cite{transformer} to perform global fusion across modalities. By alternately treating each modality as the query while the other serves as the key and value, we achieve a comprehensive cross-modal interaction that captures both visual and textual context. This global fusion mechanism ensures that the features from each modality are enriched with complementary information from the other, resulting in a more cohesive and semantically aligned representation:

\begin{align}
     \hat{Q} &= \operatorname{MHCA}(Q, V, V), \label{eq:first cross atten}\\
    \hat{V} &= \operatorname{MHCA}(V, Q, Q).
\end{align}

To capture intra-modal semantics, we utilize two multi-head self-attention (MHSA) \cite{transformer} layers on $\hat{Q}$ and $\hat{V}$, enabling a refined local fusion within each modality. This approach facilitates intricate intra-modal relationships, producing enriched representations $\tilde{Q}$ and $X$ that capture subtle contextual dependencies:
\begin{align}
     \Tilde{Q} &= \operatorname{MHSA}(\hat{Q}, \hat{Q}, \hat{Q}),\\
    X &= \operatorname{MHSA}(\hat{V}, \hat{V}, \hat{V}).
\end{align}
We input $X$ to later process. We apply the max-pooling function to obtain the sentence representation $Q^s$:
\begin{equation}
    Q^s = \operatorname{MaxPool}(\Tilde{Q}).
\end{equation}

\subsection{Temporal Consistency Learning (TCL)}
\label{sec:Temporal Consistency Learning}
 After obtaining the video-text features, we process the video-text features through both frame-level and
segment-level TCL paths to perform video-text alignment
at complementary granularities.

\paragraph{Frame-level Temporal Consistency Learning.}
\label{sec:Frame-level Temporal Consistency Learning}

We calculate the cosine similarity of the video frames and the sentence representation to extract clip-level saliency scores $Y_f$:
\begin{equation}
\label{eq:frame-level similarities}
    Y_f = \mathcal{S}(X, Q^s),
\end{equation}
where $\mathcal{S}$ is the cosine similarity function. We use the Gaussian distribution to model the temporal  information:
\begin{equation}
    G_{i}=\frac{1}{\sqrt{2 \pi} \sigma} \exp \left(-\frac{\left(\left(i-t_p\right) \cdot \frac{2}{L_{v}-1}\right)^{2}}{2 \sigma^{2}}\right),
    \label{eq:gaussian guidance}
\end{equation}
where $i$ is the $i$-th frame, $t_p$ is the index of the supervised frame, and $\sigma$ is a hyperparameter to control the variance of the Gaussian distribution. 

To capture the semantic and temporal information of the fused features from the frame-level TCL, we introduce a soft cross entropy loss to encourage the embedding of relevant video frames to be close to the language query:
\begin{equation}
    \mathcal{L_{\operatorname{frame}}} = -\sum_{i=1}^{L_v} G_i\operatorname{log}\frac{Y_f^i}{\tau},
\end{equation}
where $\tau$ is the temperature parameter and $Y_f^i$ is the clip-level saliency score of $i$-th frame.
\vspace{-1em}
\paragraph{Segment-level Temporal Consistency Learning.}
\label{sec:Segment-level Temporal Consistency Learning}

We first use sliding windows to generate proposal candidates $P=\{p_1, p_2,\cdots ,p_N\}$, where $N$ denotes the number of proposal candidates. Then, we use max-pooling to extract the proposal feature $p^f_i=\operatorname{MaxPool}(p_i) \in \mathbb{R}^d$. Thus we get the features of proposal candidates $P^f=\{p^f_1, p^f_2,\cdots ,p^f_N\}$. We calculate the cosine similarity between the proposals and the sentence representation to extract segment-level semantic scores $Y_s$:
\begin{equation}
\label{eq:proposal-level similarities}
    Y_s = \mathcal{S}(P^f, Q^s),
\end{equation}
where $\mathcal{S}$ is the cosine similarity function. We design a hierarchical contrastive alignment loss strategy to optimize the segment-level TCL, detailed in Sec.~\ref{sec:Hierarchical Contrastive Alignment Loss}.

\subsection{Cross-Consistency Guidance}
\label{sec:cross consistency guidance}

\qi{To enhance video-text alignment, we leverage the synergy between frame-level saliency detection and segment-level moment localization.
Specifically, we introduce a cross-consistency guidance mechanism, which integrates frame-level consistency guidance (FCG) and segment-level consistency guidance (SCG), where the output of one TCL is used to generate a mask to refine the other.}

\paragraph{Frame-Level Consistency Guidance.}
\label{sec:Frame Level Guidance}
We design a frame-level consistency guidance module to investigate the influence of the frame-level TCL on the segment-level TCL. We focus on how to utilize the frame-level saliency scores to enhance the fine-grained alignment of video and text in the segment-level TCL. The implementation procedure is illustrated in Fig.~\ref{fig:model}~(b).

The results of frame-level TCL $Y_f$ are sent to the saliency-aware mask generator. We first apply softmax on $Y_f$, followed by a max normalization to obtain the saliency-aware consistency mask $M_f$:
\begin{equation}
    M_f = \operatorname{maxNorm}(\operatorname{softmax}((Y_f)).
\end{equation}
The obtained $M_f$ integrates the saliency information of video frames and can provide guidance for segment-level TCL. In this manner, we enhance the video-sentence representations with element-wise product:
\begin{equation}
\label{eq:frame-level consistency guidance}
    X_s = X + X \odot M_f.
\end{equation}
Then, we feed the enhanced features $X_s$ into the segment-level TCL again to get the final segment proposal scores, which are sent into hierarchical contrastive alignment loss for optimization.

\paragraph{Segment-Level Consistency Guidance.}
\label{sec:Segment Level Guidance}
Similar to the frame-level guidance, we introduce the segment-level consistency guidance with the semantic-aware mask generator to explore the influence of the segment-level TCL on the frame-level TCL. As illustrated in Fig.~\ref{fig:model}~(c), we input the results of the segment-level TCL into the semantic-aware mask generator to generate semantic-aware consistency masks $M_s$. We first select the top-$k$ proposals based on the cosine similarity with the sentence feature. 
Then, we feed the selected proposal into the binary vector mapping module.
Here, we construct a vector with a length of $L_v$ for each proposal in the following way: If the index of the vector lies within the corresponding proposal interval, we set the value at that position to $1$; otherwise, we set it to $0$. 
Through this binary vector mapping step, we can obtain $k$ binary vectors. 
Then, we sum these $k$ binary vectors and apply softmax normalization to get the semantic-aware consistency mask $M_s$. The obtained $M_s$ integrates the prediction results from $k$ positive moments and provide guidance for frame-level TCL. Then we enhance the input of the frame-level TCL with element-wise product using $M_s$:
\begin{equation}
\label{eq:proposal-level consistency guidance}
    X_f = X + X \odot M_s.
\end{equation}
Next, we feed the enhanced features $X_f$ into the frame-level TCL again to get the final frame saliency scores, which are sent into $\mathcal{L}_{\operatorname{frame}}$ for optimization.

\subsection{Hierarchical Contrastive Alignment Loss}
\label{sec:Hierarchical Contrastive Alignment Loss}
The objective of this loss is to learn a multi-modal embedding space, where the query sentence feature should be well aligned with the feature of the corresponding target moment and far away from irrelevant video moments. For each moment, we calculate the Gaussian weight with start position s, middle postion, and end position t:
\begin{equation}
    w_{i} = \frac{1}{3}(G(s) + G(t) + G(\frac{s+t}{2})).
\end{equation}
Motivated by \citet{d3g}, we consider the new prior $h$ based on both the Gaussian weights and semantic consistency:
\begin{equation}
    h_{i} = w_{i} \cdot Y_f^i.
\end{equation}
Then we sample top-$k$ candidate moments from $P^f$ based on the new prior $h$, denoted as $P^{+}=\{p^f_1,p^f_2.\cdots,p_k^f\} \in \mathbb{R}^{k\times d}$. We also sample the corresponding Gaussian weights, which are denoted as $W^p=\{w_1,w_2.\cdots,w_k\} \in \mathbb{R}^{k}$. Different from D3G~\cite{d3g}, we consider the negative video moments hierarchically. Specifically, we design the intra-video selective alignment loss and inter-video contrastive mining loss, as shown in Fig.~\ref{fig:model}~(d). For intra-video selective alignment loss, the negative candidate moment is the moment that does not contain the annotated frame in the video, which is denoted as $P^-$.  
It is based on the InfoNCE loss~\cite{infonce}, which enforces the similarities between the language query and a group of positive candidate moments while pushing away negative pairs in a joint embedding space:
\begin{equation}
\resizebox{\linewidth}{!}{%
$
\begin{aligned}
\mathcal{L}_{\text{intra}} &= -\frac{1}{k} \sum_{p_{i}^f \in P^{+}} w_i \log \frac{\exp \left(S\left(p_{i}^f, Q^{s}\right) / \tau\right)}{\text{SUM}}, \\
\text{where SUM} &= \sum_{p_{i}^f \in P^{+}} \exp(S(p_{i}^f, Q^{s})/\tau) + \sum_{p_{j}^f \in P^{-}} \exp(S(p_{j}^f, Q^{s})/\tau),
\end{aligned}
$
}
\end{equation}
here, $S(\cdot)$ is the cosine similarity function, $\tau$ is the temperature parameter and $k$ is the number of positive moments.

The inter-video contrastive
mining loss is also an InfoNCE loss calculated in a mini-batch. 

Within a mini-batch, we treat all positive candidates from matched video-query pairs as positive terms, while considering all moments from unmatched video-query pairs as negative terms:
\begin{equation}
\resizebox{\linewidth}{!}{%
$
\begin{aligned}
    \mathcal{L}_{\text{inter}} &= -\frac{1}{k \times B} \sum_{b=0}^{B} \sum_{p^f_{b,i} \in P_b^{+}} w_i \log \frac{\exp \left( S \left( p^f_{b,i}, Q_b^s \right) / \tau \right)}{\sum_{p^f_{b,i} \in P_b^{+}} \exp \left( S \left( p^f_{b,i}, Q_b^s \right) / \tau \right) + \mathcal{N}}, \\
    \text{where}\ \mathcal{N} &= \sum_{j \neq b} \left( \exp \left( S \left( p^f_{b,i}, Q_j^s \right) / \tau \right) + \exp \left( S \left( p^f_j, Q_b^s \right) / \tau \right) \right),
\end{aligned}
$
}
\end{equation}
here, $p^f_{b,i}$ and $Q^s_b$ denote the representation of \textit{i-th} positive candidates of $b$-th video and the sentence representation of the $b$-th query in a mini-batch. $k$ is the size of $P_b^{+}$, which denotes the set of positive candidates of $b$-th video. $B$ is batch size. $\mathcal{N}$ denotes the negative terms of none paired video-sentence in the mini-batch. For the $i$-th positive proposal from the $b$-th video ($p^f_{b,i}$), $\mathcal{N}$ include: 1) pairs between $p^f_{b,i}$ and queries from other videos ($Q^s_j, j\neq b$), and 2) pairs between proposals from other videos ($p^f_j, j\neq b$) and the current video's query ($Q^s_b$).

\subsection{Training and Inference}
\label{sec:training and inference}

\paragraph{Training.} 
For the optimization of frame-level temporal consistency learning, the training loss is $\mathcal{L_{\operatorname{frame}}}$. For the optimization of segment-level temporal consistency learning, the training loss is $\mathcal{L}_{\text {inter }}$ and $\mathcal{L}_{\text {intra}}$. We also include the KL divergence loss following ViGA~\cite{viga}, which is a KL divergence loss between the attention distribution of our cross attention layer (Eq.~\ref{eq:first cross atten}) and the Gaussian guidance (Eq.~\ref{eq:gaussian guidance}):
\begin{equation}
    \mathcal{L}_{\operatorname{KL}} =  D_{\operatorname{KL}}(G \parallel A), 
\end{equation}
where $A$ is the sentence-level attention distribution in cross-attention layer and $D_{\text{KL}}$ is the Kullback-Leibler~(KL) divergence.
Overall, the training objective of our model is:
\begin{equation}
    \mathcal{L}_{total} = \mathcal{L_{\operatorname{frame}}} + \lambda_{\text{intra}} \mathcal{L}_{\text {intra }} + \lambda_{\text{inter}}\mathcal{L}_{\text {inter}} + \lambda_{\text{KL}}\mathcal{L}_{\operatorname{\text{KL}}}. 
\label{eq:loss_sum}
\end{equation}

\paragraph{Inference.} 
During inference, we combine the final proposal scores after the cross-consistency guidance with an attention-guided strategy. First, we identify an anchor frame through the maximum attention value from the cross-modal attention distribution $A$. Then, we retain only those proposals that contain this anchor. We select the one with the highest proposal score as the final prediction among the remaining proposals. This strategy effectively handles the absence of annotated frames during testing while maintaining consistency with our training approach.

\section{Experiments}

\subsection{Datasets}

We evaluate our method on two widely adopted benchmark datasets: Charades-STA~\cite{CTRL} with short-duration videos and TACoS~\cite{tacos} with long-duration videos.

\noindent
\textbf{Charades-STA}  is built on the Charades dataset~\cite{charades} and transformed into NLVL task by \citet{CTRL}. It consists of 16,128 video-sentence pairs, with 12,408 pairs used for training and 3,720 for testing. The average video length, moment length, and query length are 30.60 seconds, 8.09 seconds, and 7.22 words, respectively.

\noindent
\textbf{TACoS} is a dataset comprising 127 videos derived from the MPII Cooking Composite Activities video corpus~\cite{rohrbach2012script}, focusing on detailed cooking activities in a kitchen environment. We adhere to the standard split as provided by~\cite{tacos}, which includes 10,146, 4,589, and 4,083 moment-sentence pairs for training, validation, and testing, respectively. To ensure a fair comparison with existing methods, we report the evaluation results on the test set.

In particular, we replace the boundary supervision information in the original dataset with point-supervised information, which is derived from a randomly selected point within the boundary as in~\cite{viga}.

\vspace{-0.5em}
\subsection{Experimental Settings}
\paragraph{Evaluation Metrics.}
We use the two widey used metrics as in~\cite{viga}: (1)~``$Recall@n, IoU=m$'', which means the percentage of at least one of the top-n proposals having Intersection over Union (IoU) with the target moment larger than $m$, and (2)~mean IoU (mIoU), which is the average IoU over all the test samples. We report the results with $n=1$ and $m\in\{0.3,0.5,0.7\}$.

\begin{table}[t]
\setlength{\abovecaptionskip}{0.1cm}
\setlength{\belowcaptionskip}{0.1cm}
\centering

\resizebox{\linewidth}{!}{
\begin{tabular}{l|l|c|c|c|c}
\toprule
\textbf{Type} & \textbf{Method} & \textbf{R1@0.3} & \textbf{R1@0.5} & \textbf{R1@0.7} & \textbf{mIoU} \\
\midrule
\multirow{4}{*}{FS} & 2D-TAN~\cite{2D-TAN}\textit{\footnotesize AAAI'20} & - & 50.62 & 28.71 & - \\
 & SS~\cite{ss}\textit{\footnotesize ICCV'21} & - & 56.97 & 32.74 & - \\
 & FVMR~\cite{FVMR}\textit{\footnotesize ICCV'21} & - & 55.01 & 33.74 & - \\
 & ADPN~\cite{adpn}\textit{\footnotesize ACMMM'21} & 70.35 & 55.32 & 37.47 & 51.15 \\
 & QD-DETR~\cite{qddetr}\textit{\footnotesize CVPR'23} & - & 50.67 & 31.02 & - \\
\midrule
\multirow{4}{*}{WS} & CWG~\cite{cwg}\textit{\footnotesize AAAI'22} & 43.41 & 31.02 & 16.53 & - \\
 & CPL~\cite{CPL}\textit{\footnotesize CVPR'22} & 66.40 & 49.24 & 22.39 & - \\
 & IRON~\cite{IRON}\textit{\footnotesize CVPR'23} & 70.28 & 51.33 & 24.31 & - \\
 & PPS~\cite{PPS}\textit{\footnotesize AAAI'24} & 69.06 & 51.49 & 26.16 & - \\
 & DSRN~\cite{dsrn}\textit{\footnotesize TMM'25} & 67.95 & 52.18 & 24.60 & 45.48\\
\midrule
\multirow{4}{*}{PS} & ViGA (baseline)~\cite{viga}\textit{\footnotesize SIGIR'22} & 71.21 & 45.05 & 20.27 & 44.57 \\
 & PSVTG~\cite{psvtg}\textit{\footnotesize TMM'22} & 60.40 & 39.22 & 20.17 & 39.77 \\
 & CFMR~\cite{cfmr}\textit{\footnotesize ACMMM'23} & - & 48.14 & 22.58 & - \\
 & D3G~\cite{d3g}\textit{\footnotesize ICCV'23} & - & 43.82 & 20.46 & - \\
 & SG-SCI~\cite{sgsci}\textit{\footnotesize ACMMM'24} & 70.30 & 52.07 & 27.23 & 46.77 \\
 & COTEL (Ours) & \textbf{71.45} & \textbf{53.52} & \textbf{27.37} & \textbf{47.62} \\
\bottomrule
\end{tabular}}
\caption{Results on Charades-STA under fully-supervised (FS), weakly-supervised (WS), and point-supervised settings (PS). The best results for each metric are highlighted in bold.}
\vspace{-1.5em}
\label{tab:charades_stacomparison_results}
\end{table}

\noindent\textbf{Implementation Details.}
In our work, we employ the pre-trained I3D~\cite{I3D} and C3D~\cite{C3D} network to extract visual features for Charades-STA~\cite{charades} and TACoS~\cite{tacos} respectively. Following ViGA \cite{viga}, we utilize the 840B GLoVe~\cite{glove} to extract the text feature. The model dimension $d$ is set to 512, and we employ the AdamW~\cite{adamw} with a learning rate of 1e-4, which decays by half when reaching a plateau for 3 epochs during training. We set $\tau=0.5$ for Charades-STA and 0.1 for TACoS. We set $k=15$ for Charades-STA and 20 for TACoS. By default, we set  $\lambda_{\text{KL}} = 1$ for both datasets. We set $\lambda_{\text{intra}} = 0.5$, $\lambda_{\text{inter}}=0.05$  for Charades-STA. For TACoS, we set $\lambda_{\text{intra}} = 0.1$, $\lambda_{\text{inter}}=1$. The batch size for the Charades-STA and TACoS are set to 256 and 64, respectively. All experiments are conducted on an NVIDIA GeForce RTX 4090 with 24 GB memory.

\subsection{Comparison with State-of-the-art Methods}
\begin{table}[t]
\setlength{\abovecaptionskip}{0.1cm}
\setlength{\belowcaptionskip}{0.1cm}
\centering

\resizebox{\linewidth}{!}{
\begin{tabular}{l|l|c|c|c|c}
\toprule
\textbf{Type} & \textbf{Method} & \textbf{R1@0.3} & \textbf{R1@0.5} & \textbf{R1@0.7} & \textbf{mIoU} \\
\midrule
\multirow{4}{*}{FS} & 2D-TAN~\cite{2dtan}\textit{\footnotesize AAAI'20} & 37.29 & 25.32 & - & - \\
 & SS~\cite{ss}\textit{\footnotesize ICCV'21} & 41.33 & 29.56 & - & - \\
 & FVMR~\cite{FVMR}\textit{\footnotesize ICCV'21} & 41.48 & 29.12 & - & - \\
 & MS-DETR~\cite{MS-DETR}\textit{\footnotesize ACL'23} & 47.66 & 37.36 & 25.81 & 35.09 \\
 & MomentDiff~\cite{momentdiff}\textit{\footnotesize NIPS'23} & 44.78 & 33.68 & - & - \\
\midrule
\multirow{3}{*}{WS} & SCN~\cite{scn}\textit{\footnotesize AAAI'20} & 11.72 & 4.75 & - & - \\
 & CNM~\cite{cnm}\textit{\footnotesize AAAI'22} & 7.20 & 2.20 & - & - \\
 & CPL~\cite{CPL}\textit{\footnotesize CVPR'22} & 11.42 & 4.12 & - & - \\
 & SiamGTR~\cite{SiamGTR}\textit{\footnotesize CVPR'24} & 26.22 & 10.53 & - & 20.48 \\
\midrule
\multirow{5}{*}{PS} & ViGA (baseline)~\cite{viga}\textit{\footnotesize SIGIR'22} & 19.62 & 8.85 & 3.22 & 15.47 \\
 & PSVTG~\cite{psvtg}\textit{\footnotesize TMM'22} & 26.34 & 10.00 & 3.35 & 17.39 \\
 & CFMR~\cite{cfmr}\textit{\footnotesize ACMMM'23} & 25.44 & 12.82 & - & - \\
 & D3G~\cite{d3g}\textit{\footnotesize ICCV'23} & 27.27 & 12.67 & 4.70 & - \\
 & SG-SCI~\cite{sgsci}\textit{\footnotesize ACMMM'24} & 37.47 & 20.59 & 8.27 & 23.83 \\
 & COTEL (Ours) & \textbf{38.39} & \textbf{22.79} & \textbf{8.74} & \textbf{25.87}\\
\bottomrule
\end{tabular}}
\caption{Results on TACoS under fully-supervised (FS), weakly-supervised (WS), and point-supervised settings (PS). Bold fonts denote the best result in point-supervised method.}
\label{tab:tacos_supervision_methods}
\end{table}

We compare the proposed method with several state-of-the-art methods at different levels of supervision, including fully-supervised methods (2D-TAN~\cite{2D-TAN}, SS~\cite{ss}, FVMR~\cite{FVMR}, MS-DETR~\cite{MS-DETR}, MomentDiff~\cite{momentdiff}, ADPN~\cite{adpn}, QD-DETR~\cite{qd_detr}), weakly-supervised methods (SCN~\cite{scn}, CNM~\cite{cnm}, IRON~\cite{IRON}, CPL~\cite{CPL}, DSRN~\cite{dsrn}, CWG~\cite{cwg}), and point-supervised methods (ViGA~\cite{viga}, SG-SCI~\cite{sgsci}, PSVTG~\cite{psvtg}, D3G~\cite{d3g}, CFMR~\cite{cfmr}).
\begin{table}[t]
\setlength{\abovecaptionskip}{0.1cm}
\setlength{\belowcaptionskip}{0.1cm}
\centering

\resizebox{0.8\linewidth}{!}
{
\begin{tabular}{l|cccc}
\toprule
Method & R1@0.3 & R1@0.5 & R1@0.7 & mIoU \\
\midrule
Frame-level TCL & 69.74 & 46.14 & 21.65 & 44.21 \\
Segment-level TCL & 70.04 & 47.77 & 23.89 & 45.46 \\
Both & 70.12 & 48.45 & 25.26 & 46.78 \\
\bottomrule
\end{tabular}
}
\caption{Ablation study on individual paths without cross-consistency guidance.}
\label{tab:ablation_paths}
\vspace{-1em}
\end{table}
The results are  shown in Tab.~\ref{tab:charades_stacomparison_results} and Tab.~\ref{tab:tacos_supervision_methods}.
The results show that:
\begin{itemize}
    \item Our proposed COTEL method achieves favorable performance among all point-supervised approaches, demonstrating the effectiveness of collaborative temporal consistency learning for video-text alignment. Compared to other point-supervised methods, our method consistently outperforms across all metrics. Compared to ViGA~\cite{viga}, we achieve obvious gains 18.77\% and 13.94\% at R1@0.3 and R1@0.5 on TACoS, respectively. On Charades-STA, we outperform ViGA~\cite{viga} by 8.47\% and 7.1\%  at R1@0.5 and R1@0.7, respectively.
    \item Our proposed COTEL method outperforms weakly-supervised approaches, demonstrating the effectiveness of point-level supervision with trivial increment of annotation cost.  On Charades-STA (Tab.~\ref{tab:charades_stacomparison_results}), COTEL achieves an \textit{R1@0.5} of 53.52, surpassing IRON's 51.33, with a larger margin at \textit{R1@0.7} (27.37 vs. 24.31). On the more challenging TACoS dataset (Tab.~\ref{tab:tacos_supervision_methods}), COTEL's advantage is even more evident, achieving an \textit{R1@0.5} of 22.79 compared to SiamGTR's 10.53. 

    \item Our COTEL narrows the performance gap between fully-supervised and point-supervised approaches. Specifically, we outperform 2D-TAN~\cite{2D-TAN} by 2.9\% on Charades-STA in terms of R1@0.5. In TACoS, we outperform 2D-TAN~\cite{2D-TAN} by 1.1\%  in terms of R1@0.3. 
\end{itemize}

\subsection{Ablation Study}

\begin{table}[t]
\setlength{\abovecaptionskip}{0.1cm}
\setlength{\belowcaptionskip}{0.1cm}
\centering
\resizebox{0.7\linewidth}{!}
{
\begin{tabular}{cccccc}
\toprule
FCG & SCG & R1@0.3 & R1@0.5 & R1@0.7 & mIoU \\
\midrule
    &   & 70.12 & 48.45 & 25.26 & 46.78 \\
\checkmark &  &  70.56 & 50.24 & 26.83 & 47.02 \\
 & \checkmark & 70.30 & 49.66 & 25.78 & 46.91 \\
 \checkmark & \checkmark & 71.45 & 53.52 & 27.37 & 47.62  \\
\bottomrule
\end{tabular}}
\caption{Ablation studies of different components in cross-consistency guidance.
}
\label{tab:components_ablation}
\end{table}

\begin{table}[t]
\setlength{\abovecaptionskip}{0.1cm}
\setlength{\belowcaptionskip}{0.1cm}
\centering

\resizebox{\linewidth}{!}{
\begin{tabular}{cccccccc}
\toprule
$\mathcal{L}_{\text{KL}}$ & $\mathcal{L}_{\text{frame}}$ & $\mathcal{L}_{\text{intra}}$ & $\mathcal{L}_{\text{inter}}$ & R1@0.3 & R1@0.5 & R1@0.7 & mIoU \\
\midrule
  \checkmark&  &  &  & 61.13 & 32.18 & 11.80 & 37.02 \\
\checkmark &  \checkmark &  &  & 70.28 & 47.28 & 24.81 & 44.76 \\
\checkmark & \checkmark & \checkmark &  & 70.35 & 53.01 & 26.53 & 46.75 \\
\checkmark & \checkmark & & \checkmark & 71.31 & 49.11 & 25.42 & 46.28 \\
\checkmark & \checkmark & \checkmark & \checkmark & 71.45 & 53.52 & 27.37 & 47.62   \\
\bottomrule
\end{tabular}}
\caption{Ablation studies of different loss on  Charades-STA. }
\vspace{-1.5em}
\label{tab:loss_ablation}
\end{table}

In this section, we conduct ablation studies to validate the effectiveness of different components and the impact of hyper-parameters on Charades-STA.

\vspace{-0.5cm}
\paragraph{Ablation on Different Paths.}
We conduct an ablation study to evaluate the effectiveness of each TCL without cross-consistency guidance (CCG). As shown in Tab.~\ref{tab:ablation_paths}, combining both paths improves the performance over the single path. These results demonstrate that both paths contribute to the overall performance:
frame-level alignment for frame saliency detection and segment-level alignment for moment localization.
\vspace{-1.4em}
\paragraph{Ablation of Cross-Consistency Guidance.}
Tab.~\ref{tab:components_ablation} presents our ablation study on cross-consistency guidance components. The results demonstrate that frame-level consistency
guidance (FCG) and segment-level consistency guidance (SCG) work synergistically to achieve substantial improvements compared to using either component alone. FCG enhances performance by transforming frame-level saliency scores into consistency masks that guide segment-level TCL, particularly improving fine-grained localization metrics (R1@0.5 increases by 1.79\% and R1@0.7 by 1.57\%). 
Meanwhile, SCG functions by leveraging proposal scores to generate semantic-aware consistency masks that refine frame-level features, providing broader contextual understanding that improves performance across all metrics. 
When both mechanisms are adopted, the performance is further improved, achieving an R1@0.5 of 53.52 higher than using either component alone.
These results validate our key insight that leveraging the synergy between saliency detection and moment localization significantly improves performance.

\vspace{-1.5em}
\paragraph{Effectiveness of Different Losses.}
The final loss function comprises four components: intra-video selective alignment loss $\mathcal{L_{\operatorname{intra}}}$, inter-video contrastive mining loss $\mathcal{L_{\operatorname{inter}}}$, frame-level saliency loss $\mathcal{L_{\operatorname{frame}}}$, and KL divergence loss $\mathcal{L_{\operatorname{KL}}}$. To evaluate the effectiveness of these components, we perform an ablation study as shown in Tab.~\ref{tab:loss_ablation}.
Training with $\mathcal{L}_{\operatorname{KL}}$ alone yields inferior results, as it only enforces attention weights to match a predefined Gaussian distribution without explicit video-text alignment in semantic space. 
Adding $\mathcal{L}_{\operatorname{frame}}$ substantially improves performance by introducing basic temporal supervision. 
Building upon this base configuration, incorporating $\mathcal{L}_{\operatorname{intra}}$ further brings gains in precise temporal localization (R1@0.5 improving by {5.73\%}) by enhancing intra-video semantic discrimination, while $\mathcal{L}_{\operatorname{inter}}$ shows consistent improvements through cross-video contrastive learning. The full model achieves SOTA performance, confirming both loss components contribute complementarily. 
This hierarchical loss design enables comprehensive alignment and captures temporal relationships both within and across videos.

\begin{figure}
    \centering
    \includegraphics[width=0.75\linewidth]{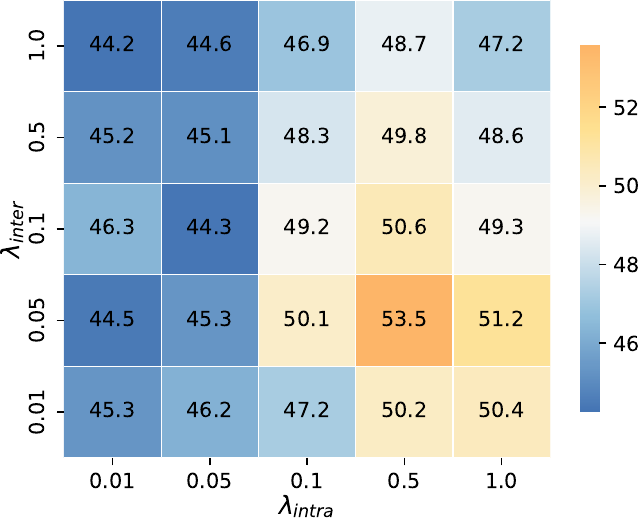}
    \caption{The effect of coefficients for intra-video selective alignment loss and inter-video contrastive mining loss on model performance. We present results of \textit{R1@0.5} on the Charades-STA.}
    \label{fig:ablation_loss_coefficients}
    \vspace{-1em}
\end{figure}

\paragraph{Ablation on the Loss Coefficients.}
In Eq.~\ref{eq:loss_sum}, $\lambda_{intra}$ and $\lambda_{inter}$ represent the importance of intra-video selective alignment loss and inter-video contrastive mining loss in the final loss function. We test different combinations of $\lambda_{intra}$ and $\lambda_{inter}$ in the range $\{0.01, 0.05, 0.1, 0.5, 1\}$, and illustrate the impact on $R1@0.5$, as shown in Fig.~\ref{fig:ablation_loss_coefficients}. The model's performance shows an overall trend of first increasing and then decreasing, with the highest performance achieved at $\lambda_{intra}=0.5, \lambda_{inter} = 0.05$.  The experimental results indicate that appropriately balancing the weights of intra-video selective alignment loss and inter-video contrastive mining loss is crucial for enhancing fine-grained boundary perception, as these components jointly improve the model's ability to differentiate between positive and negative samples within and across videos.

\subsection{Qualitative Results}

\begin{figure}
    \centering
    \includegraphics[width=\linewidth]{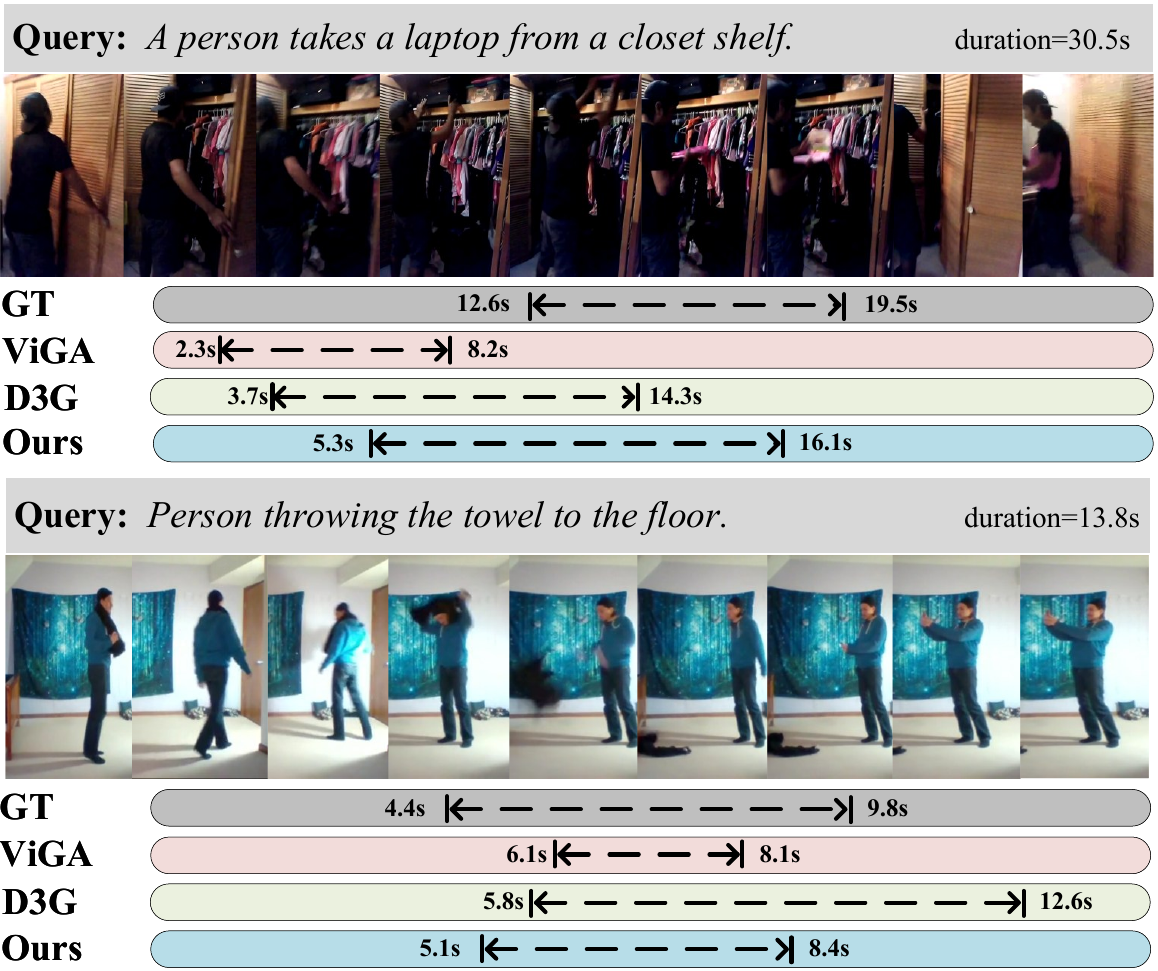}
    \caption{
    Qualitative examples of top-1 predictions on Charades-STA. We compare our method with ViGA~\cite{viga} and D3G~\cite{d3g}. GT indicates the ground truth temporal boundary.
    }
    \vspace{-1em}
    \label{fig:qualitative}
\end{figure}

We visualize some qualitative examples from the test split of the Charades-STA dataset. As shown in Fig.~\ref{fig:qualitative}, our proposed COTEL achieves more precise localization of target moments compared to existing methods. Specifically, previous methods often struggle with aligning visual content and semantic information, leading to confusion by irrelevant content. In contrast, our collaborative consistency learning framework leverages the cross-consistency guidance to learn the synergy between saliency detection and moment localization to ensure better alignment between video and query. More qualitative results are in the supplementary.

\section{Conclusion}
We propose a new collaborative consistency learning framework for the natural language video localization task under point-supervised settings.
\qi{Our \textbf{CO}llaborative \textbf{T}emporal consist\textbf{E}ncy \textbf{L}earning (COTEL) framework mitigates the challenge of limited frame annotations by leveraging frame-level and segment-level temporal consistency learning. 
The COTEL framework, combined with the Gaussian prior, facilitates semantic alignment between frame saliency and intra- and inter-sentence-moment pairs. 
Meanwhile, cross-consistency guidance enhances temporal consistency by enabling mutual learning between the two paths.}
Extensive experiments on two datasets show that our COTEL achieves state-of-the-art performance, striking a balance between localization accuracy and annotation cost. 
This work shows the potential of collaborative consistency learning to advance video understanding with limited supervision.

{
    \small
    \bibliographystyle{ieeenat_fullname}
    \bibliography{main}
}

\end{document}